# Mapping Scrambled Korean Sentences into English Using Synchronous TAGs

**Hyun S. Park**
Computer Laboratory
University of Cambridge
Cambridge, CB2 3QG, U.K.
Hyun.Park@cl.cam.ac.uk

## Abstract

Synchronous Tree Adjoining Grammars can be used for Machine Translation. However, translating a free order language such as Korean to English is complicated. I present a mechanism to translate scrambled Korean sentences into English by combining the concepts of Multi-Component TAGs (MC-TAGs) and Synchronous TAGs (STAGs).

## 1 Motivation

Tree Adjoining Grammars (TAGs) were first developed by Joshi, Levy, and Takahashi (Joshi et al., 1975). There are other variants of TAGs such as STAGs (Shieber and Schabes, 1990), and MC-TAGs (Weir, 1988). STAGs in particular can be used for machine translation and were applied to Korean-English machine translation in a military message domain (Palmer et al., 1995).

Park (Park, 1995) suggested a way of handling Korean scrambling using MC-TAGs together with a *priority* concept. However, as scrambled argument structures in Korean were represented as sets using MC-TAGs, a mechanism to combine MC-TAGs and STAGs was necessary to translate Korean scrambled sentences into English.

## 2 Korean-English Machine Translation Using STAGs

STAGs are a variant of TAGs introduced to characterize correspondences between tree adjoining languages. They can be used to relate TAGs for two different languages for machine translation (Abeillé et al., 1990). The translation process consists of three steps. The source sentence is parsed according to the source grammar. Each elementary tree in the derivation is considered with the features given from the derivation through unification. Second, the source derivation tree is transferred to a target derivation. This step maps each elementary tree in the source derivation tree to a tree in the target derivation tree by looking in the transfer lexicon. And finally, the target sentence is generated from the target derivation tree obtained in the previous step.

The transfer lexicon consists of pairs of trees, one from the source language and the other from the target language. Within the pair of trees, nodes may be linked. Whenever adjunction or substitution is performed on a linked node in a source tree, the corresponding operation applies to the linked node in the target tree.

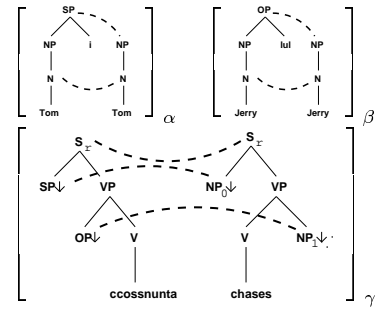

Figure 1: The K-E Transfer Lexicon

Canonical ordering of the arguments of transitive verbs in Korean is SOV. Whereas the case marker in English is implicit in the word, case markers are explicit in Korean. This is reflected in the transfer lexicon of Figure 1. So, the pair $\alpha$ in Figure 1 shows that Korean has an explicit subject case marker *i*, and the pair $\beta$ shows that Korean has an explicit object case marker *lul*. Also, the pair $\gamma$ shows the links between SOV structure of Korean to SVO structure of English.

1  K:  *Tom-i*     *Jerry-lul*   *ccossnunta.*
    Tom-NOM   Jerry-ACC   chase
   E:  Tom         chases         Jerry.

To translate sentence (1), we start with the pair $\gamma$ in Figure 1, and we substitute the pair $\alpha$ on the link from the Korean node SP to the English node NP. Then, pair $\beta$ is substituted into the NP-OP pairs in $\gamma$, thus correctly transferring sentence (1).

## 3 Handling of Scrambling in Korean Using MC-TAGs

TAGs and related formalisms, due to the extended domain of locality, can combine a lexical head and all of its arguments in a single elementary structure of the grammar. However, Becker and Rambow show that TAGs that obey the co-occurrence constraint cannot handle the full range of scrambled sentences (Becker and Rambow, 1990). As a result, non-local MC-TAG-DL (Multi-Component TAG with Dominance Link) was proposed as a way of handling scrambling[1]. Later, by adding a *priority* concept to MC-TAG-DL, Park (Park, 1995) suggested a way of handling scrambling in Korean.

### 3.1 $\alpha \mathcal{ARG}$ & $\beta \mathcal{ARG}$ structures

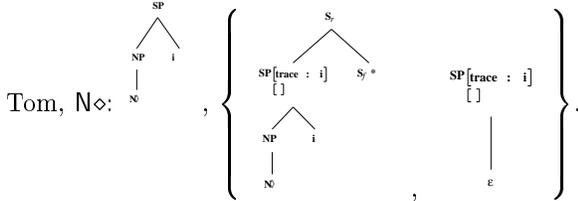

For handling scrambling, the multi-adjunction concept in MC-TAGs can be used for combining a scrambled argument and its landing site. For example, a subject (e.g., *Tom*) would have two Korean structures as above. For notational convenience, call the two structures, $\alpha \mathcal{ARG}_{\mathcal{SP}}$ and $\beta \mathcal{ARG}_{\mathcal{SP}}$, respectively. In general, $\alpha \mathcal{ARG}$ represents a canonical NP structure and $\beta \mathcal{ARG}$ represents a scrambled NP structure. $\beta \mathcal{ARG}_{\mathcal{SP}}$ shows a pair of structures for representing the scrambled subject argument. Call the left structure of $\beta \mathcal{ARG}_{\mathcal{SP}}$, $\beta \mathcal{ARG}^{\mathcal{L}}_{\mathcal{SP}}$ and the right structure, $\beta \mathcal{ARG}^{\mathcal{R}}_{\mathcal{SP}}$. $\beta \mathcal{ARG}^{\mathcal{L}}_{\mathcal{SP}}$ represents a scrambled subject, and $\beta \mathcal{ARG}^{\mathcal{R}}_{\mathcal{SP}}$ is used for representing the place where the subject would have been in the canonical sentence. Similarly, $\beta \mathcal{ARG}_{\mathcal{OP}}$ denotes a pair of structures for representing a scrambled object argument.

The basic idea is that whenever an argument is not in a scrambled position, it should be substituted into an available empty slot using the $\alpha \mathcal{ARG}$ structure. The $\beta \mathcal{ARG}$ structure will be used only when the argument is in a scrambled position so that the $\alpha \mathcal{ARG}$ structure cannot be used.

### 3.2 An Example

|   | K: | Jerry-lul | Tom-i | ccossnunta. |
| --- | --- | --- | --- | --- |
| 2 |   | Jerry-ACC | Tom-NOM | chase-DECL |
|   | E: | Tom | chases | Jerry |

From the elementary trees in Figure 2, both sentences, (1) and (2) can be derived. For example, Figures 2(a), 2(b), and 2(d) can be used for sentence (1), to derive Figure 3(a). However, for sentence (2) where the order is OSV (the object argument is scrambled), Figures 2(a), 2(c), and 2(d) are used to derive Figure 3(b) ($\beta \mathcal{ARG}^{\mathcal{L}}_{\mathcal{OP}}$ is adjoined onto S, and $\beta \mathcal{ARG}^{\mathcal{R}}_{\mathcal{OP}}$ is substituted into $\mathsf{OP}_1 \downarrow$ node.). As the **trace** feature is *locally* set within each $\beta \mathcal{ARG}$ structure, two OP nodes in Figure 3(b) are co-referenced with the same variable, $< 1 >$, indicating where the object should have been in the canonical sentence.

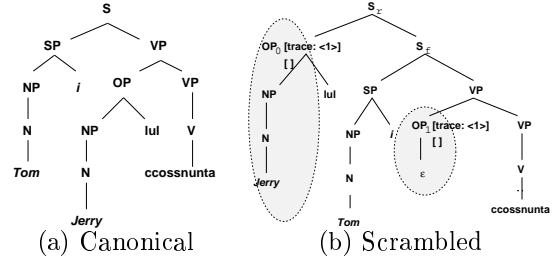

Figure 2: Elementary Trees

Figure 3: Derived Trees

Each elementary tree is given a *priority*. A higher *priority* is given to $\alpha \mathcal{ARG}$ structure over $\beta \mathcal{ARG}$. Generally, when a structure given a higher *priority* over others can be successfully used for the final derivation of a sentence, the remaining structures will not be tried at all. Only when the highest priority structure fails will the next available structure be tried[2].

## 4 Using MC-TAGs in STAGs

For mapping Korean to English, the simple object (NP) structure of English (e.g., the right structure of $\beta$ pair in Figure 1) can be mapped to two structures, i.e., $\alpha \mathcal{ARG}_{\mathcal{OP}}$ and $\beta \mathcal{ARG}_{\mathcal{OP}}$, thus generating two possible lexical pairs.

---

[1] An additional constraint system called *dominance links* was added, thus giving rise to MC-TAG-DL.

[2] As a way of implementing a verb-final condition in Korean, $\beta \mathcal{ARG}^{\mathcal{R}}_{\mathcal{SP}}$ structure is dominated by $\beta \mathcal{ARG}^{\mathcal{L}}_{\mathcal{SP}}$, and each S-type verb elementary tree will have an $\mathcal{NA}$ constraint on the root node, which guarantees that $\beta \mathcal{ARG}$ type structure cannot be adjoined onto the partially derived tree unless its predicate structure (its S-type verb elementary tree) is already part of the partial derived tree up to that point. An example including long-distance scrambling is shown in (Park, 1995).

For translating sentence (1), the $\alpha\mathcal{ARG}_{\mathcal{OP}}$–NP pair is used for *Jerry* (similar to the $\beta$ pair in Figure 1). However, in sentence (2), the $\beta\mathcal{ARG}_{\mathcal{OP}}$–NP pair should be used instead for translating the scrambled argument *Jerry* (i.e., Figure 4(a)). Thus, it is necessary that a Korean $\beta\mathcal{ARG}$ structure (MC-TAG) be mapped to an English NP structure (TAG) to transfer a scrambled argument in Korean. I assume that there is one **head structure** for each MC-TAG structure, and that the $\beta\mathcal{ARG}^{\mathcal{R}}$ (place holder structure) is the **head structure** for each $\beta\mathcal{ARG}$ structure. The root node of the **head structure** is always mapped to the root node of the target (English) structure.

Usually, the nodes in the source language should be linked to each relevant node in the target language, and vice versa (in STAGs). However, in the case that it is a multi-component structure (e.g., $\beta\mathcal{ARG}$), an adjunction node need not necessarily be linked to any node. If it is not linked to any node of the target language, the structure can be freely adjoined onto any available node of the partially derived tree of the source language, which is approximately what scrambling is about. However, substitution nodes will always be linked (the difference between a substitution node and an adjunction node is that an adjunction node does not introduce a new structure to the partially derived tree whereas a substitution node always does).

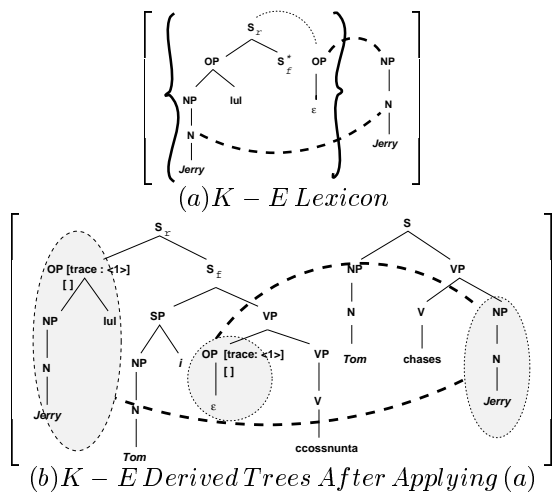

Figure 4: K-E Transfer Lexicon and Derived Tree

In Figure 4(a), the root node NP of an English TAG is mapped to the OP node of $\beta\mathcal{ARG}^{\mathcal{R}}_{\mathcal{OP}}$ of a Korean TAG which is a **head structure**. All the other nodes are mapped to each relevant node except $S^*_f$. As it is not linked, $\beta\mathcal{ARG}^{\mathcal{L}}_{\mathcal{OP}}$ can be adjoined onto any available node in the partially derived Korean tree. Actually, the restriction on whether $\beta\mathcal{ARG}^{\mathcal{L}}_{\mathcal{OP}}$ can be adjoined onto a certain node does not come from the formalism of Synchronous TAGs, but purely from the grammar of Korean TAGs. Figure 4(b) shows the final derived trees for both Korean and English after applying 4(a) to the partially derived trees.

## 5 Conclusion and Future Direction

Using MC-TAGs allows the scrambled argument structure to be represented as a *single* (set) structure. This makes possible the mapping of Korean scrambled argument structures into English argument structures. The application of similar mechanisms for other languages and for mapping *quasi logical forms* to *logical forms* (Alshawi et al., 1992) using STAGs is also being investigated.